\pdfoutput=1
\documentclass[11pt]{article}

\usepackage[final]{EMNLP2022}
\usepackage{times}
\usepackage{latexsym}

\usepackage[T1]{fontenc}
\usepackage[utf8]{inputenc}

\usepackage{microtype}

\usepackage{inconsolata}

\usepackage{anyfontsize}

\usepackage{amsmath}
\usepackage{amsfonts}
\usepackage{cleveref}
\usepackage{xcolor, color}
\usepackage{graphicx}
\usepackage{epstopdf}
\usepackage{multirow}
\usepackage{booktabs}
\usepackage{comment}
\usepackage{float}
\usepackage{enumitem}
\usepackage{makecell}
\usepackage{array}
\usepackage{courier}
\usepackage{footnote} 
\usepackage{diagbox}
\usepackage{soul}
\usepackage{longtable}
\usepackage[export]{adjustbox}
\usepackage{subcaption}
\usepackage{makecell}

\newcommand{\newparagraph}[1]{\vspace{5pt} \noindent \textbf{#1}}

\title{Probing Cross-modal Semantics Alignment Capability \\ from the Textual Perspective}

\author{
Zheng Ma\quad 
Shi Zong\quad 
Mianzhi Pan\quad 
Jianbing Zhang{\thanks{\ \ \ Corresponding author.}} \quad \\
\textbf{Shujian Huang}\quad 
\textbf{Xinyu Dai}\quad 
\textbf{Jiajun Chen}\quad \\
Nanjing University\\
{\tt \{maz, panmz\}@smail.nju.edu.cn}\\
{\tt \{szong, zjb, daixinyu, huangsj, chenjj\}@nju.edu.cn}
}

\begin{document}
\maketitle

\begin{abstract}

In recent years, vision and language pre-training (VLP) models have advanced the state-of-the-art results in a variety of cross-modal downstream tasks. 
Aligning cross-modal semantics is claimed to be one of the essential capabilities of VLP models. 
However, it still remains unclear about the inner working mechanism of alignment in VLP models. 
% the inner working mechanism of VLP models on cross-modal semantics alignment is still unclear.
In this paper, we propose a new probing method that is based on image captioning to first empirically study the cross-modal semantics alignment of VLP models.
Our probing method is built upon the fact that given an image-caption pair, the VLP models will give a score, indicating how well two modalities are aligned; maximizing such scores will generate sentences that VLP models believe are of good alignment. 
Analyzing these sentences thus will reveal in what way different modalities are aligned and how well these alignments are in VLP models.
We apply our probing method to five popular VLP models, including UNITER, ROSITA, ViLBERT, CLIP, and LXMERT, and provide a comprehensive analysis of the generated captions guided by these models. 
Our results show that VLP models 
(1) focus more on just aligning objects with visual words, while neglecting global semantics;
(2) prefer fixed sentence patterns, thus ignoring more important textual information including fluency and grammar;
and (3) deem the captions with more visual words are better aligned with images.
% (1) pay more attention to aligning objects and visual words, 
% (2) have preferences on fixed sentence patterns, 
% and (3) deem the captions containing more visual words are more consistent with images.
These findings indicate that VLP models still have weaknesses in cross-modal semantics alignment and we hope this work will draw researchers' attention to such problems when designing a new VLP model.\footnote{Our code is publicly available at \url{https://github.com/aaronma2020/probing_vlp}}
\end{abstract}

\begin{table}[h!]
\centering
\resizebox{0.48\textwidth}{!}{
\scriptsize
\begin{tabular}{cm{.24\textwidth}}
\multicolumn{2}{c}{\footnotesize \textbf{Which caption matches the image better?}} \\\midrule
% Example 1
\multirow{8}{*}{\begin{minipage}{.24\textwidth}\includegraphics[width=\linewidth]{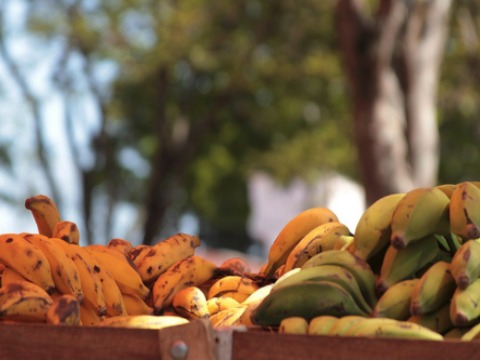}\end{minipage}} & \textbf{(a)} \emph{a bunch of bananas on a table} \\\cmidrule{2-2}
& \textbf{(b)} \emph{a bunch of the bananas table} \\\cmidrule{2-2}
& \textbf{(c)} \emph{a closeup image that is a a bunch but a a bananas but a a table}  \\\cmidrule{2-2}
& \textbf{(d)} \emph{bunch that the bananas at the table}  \\\cmidrule{2-2}
& \textbf{(e)} \emph{a image image of a bunch near a bananas near a table} \\\cmidrule{2-2}
& \textbf{(f)} \emph{a bunch of bananas of a table} \\\midrule

 \textbf{Models} &  \makecell[c]{\textbf{Choice}}\\\midrule
\textbf{UNITER} &  \footnotesize \makecell[c]{(b)} \\
\textbf{ROSITA} &  \footnotesize \makecell[c]{(c)} \\
\textbf{ViLBERT} &  \footnotesize \makecell[c]{(d)} \\
\textbf{CLIP} &  \footnotesize \makecell[c]{(e)} \\
\textbf{LXMERT} & \footnotesize \makecell[c]{(f)} \\\midrule
\textbf{Human} &  \footnotesize \makecell[c]{(a)} \\
% \textbf{} choose (c) \\\midrule
% \textbf{} (d) & \textbf{} (e) \\\midrule
% \textbf{LXMERT} (f) & \textbf{Human} (a) \\
\bottomrule
\end{tabular}
}
\caption{Given an image, VLP models are asked to choose from a reasonable caption (a) generated by a captioning model and five unreadable captions (b) - (f) modified from (a). We observe that all VLP models pick unreadable captions.}
\label{tb:intro_example}
\vspace{-.5cm}
\end{table}

\section{Introduction}

% 介绍VLP
Vision-language pre-training (VLP) models are designed to understand visual information, textual semantics, and cross-modal relationships. 
To align cross-modal semantics, most of these VLP models follow two structures: The single-stream architecture \citep{DBLP:journals/corr/abs-1909-11740, DBLP:conf/mm/CuiYWZZWY21} aligns the cross-modal information as they start being fed into the model; while alignment happens in later layers for the two-stream architecture \citep{DBLP:conf/nips/LuBPL19, DBLP:conf/icml/RadfordKHRGASAM21, DBLP:conf/emnlp/TanB19} . 
% Single-stream architecture aligns the cross-modal information as they start being fed, and two-stream architecture  aligns them in intermediate layers. 
Besides, both types of models are generally trained with a cross-modality matching task during pre-training \citep{DBLP:journals/corr/abs-1909-11740} or a contrastive loss for better alignment \citep{DBLP:conf/nips/LiSGJXH21, DBLP:journals/corr/abs-2111-07783}.
These models have achieved state-of-the-art in many cross-modal tasks, such as image-text retrieval \citep{DBLP:conf/eccv/LeeCHHH18} and  visual question answering \citep{DBLP:conf/iccv/AntolALMBZP15}.
% \shi{add cite XXXX}.

% \begin{figure}
%     \centering
%     \includegraphics[width=0.49\textwidth]{imgs/pmz.drawio.pdf}
%     \caption{Caption \shi{change}}
%     \label{fig:intro_example}
% \end{figure}

%介绍VLP probing task和存在的问题
Although VLP models have excellent performance on a large panoply of cross-modal tasks, it remains unclear which information VLP models have aligned between different modalities.
% \shi{not clear}
Some studies have investigated this problem and designed a series of probing tasks to explore it \citep{DBLP:journals/corr/abs-2012-12352, DBLP:conf/coling/LindstromBBD20}. 
However, these probing tasks almost probe VLP models based on the classification task, such as classifying or counting objects and recognizing consistencies between regions and phrases.
We argue that the simple classified probing tasks fail to explicitly explore VLP models' inner alignment mechanism, as they only require models to use partial information, such as a region, a phrase, or both, to make a decision. 
% \shi{what is global, what is partical?}.
To illustrate this point, in \Cref{tb:intro_example}, we present a reasonable caption and five unreadable captions. 
VLP models are asked which caption fits the image better. Surprisingly, no model chooses reasonable caption (a).
% \shi{what does this mean?}
It is thus reasonable to question the capability of VLP models for cross-modal alignment and motivates us to study which kind of captions VLP models deem match images better.
 
Without doubt, it is too difficult and nearly impossible to find out all unreasonable captions the models favor, by first generating some manually and then testing them, as in fact, we do not know the model's preferences beforehand.
% \shi{impossible too strong}
However, if we can use the matching score from VLP models as a signal to train a cross-modal generated model, then maximizing such scores will amplify the models' preferences and reflect them in the generated sentences.
Inspired by this idea, in this paper we present the first work of using a novel probing method to empirically study cross-modal alignment via the image captioning task. 
The generated captions from our probing method have higher matching scores, which are considered to fit images better and contain important alignment information that the VLP models focus on.

% Based on the above flow, we evaluate
We then apply our probing method to five powerful and representative VLP models, including UNITER, ROSITA, ViLBERT, CLIP, and LXMERT, which cover two mainstream architectures and two mainstream alignment tasks.
Our approach explicitly maps the alignment information to the generated captions. 
Through analyzing these captions, we discover that objects and visual words tend to receive more attention in VLP models' inner alignment mechanism, and the more nouns the captions contain, the more VLP models deem they match images. 
It indicates these VLP models are overwhelmingly dependent on partial information (objects in images and visual words in captions), rather than the whole semantics of images and captions when they judge whether image-caption pairs are aligned. 
Moreover, we find that these models favor certain sentence patterns (details in \Cref{sec:analysis}).
Captions that follow these patterns will be considered more consistent with images. It suggests when deciding whether image-caption pairs are aligned, VLP models ignore more important textual information, such as fluency and grammar.

\section{Related Work}

% Inspiring by BERT \citep{DBLP:conf/naacl/DevlinCLT19}, a larger panoply of VLP models are designed to jointly represent visual and textual features \citep{DBLP:conf/nips/LuBPL19,DBLP:journals/corr/abs-1909-11740,DBLP:conf/mm/CuiYWZZWY21}, which have achieved the state-of-the-art performances for a variety of cross-modal tasks.

% inner mechanisms of these VLP models still remain unclear. 
% Therefore, some studies begin to analyze VLP models.

There has been an increasing number of papers studying how different modalities are aligned in VLP models.
We now review recent representative studies.
\citet{DBLP:conf/eccv/CaoGCY0020} report the dominance of textual modality with VALUE, a comprehensive framework they introduce including multiple probing tasks.
\citet{DBLP:conf/coling/LindstromBBD20} design three probing tasks to analyze linguistic properties of multi-modal embeddings,
such as estimating the number of object instances in the image.
% including \textit{ObjectCategories}, \textit{NumObjects} and \textit{SemanticCongruence}. 
All these tasks employ a simple neural network classifier to probe the ability of VLP models in certain aspects or reveal the importance of textual compared with visual information. 

\citet{DBLP:journals/corr/abs-2012-12352} evaluate VLP models on count, a task requiring the model to correctly predict the number of objects in an image, and find several prevailing VLP models that fail to identify entities in an image.  
% Following this idea, we discard simple classified probing tasks used in \citet{DBLP:conf/coling/LindstromBBD20}. 
Moreover, \citet{DBLP:conf/acl/ParcalabescuCMF22} propose VALSE to test VLP models for their vision-linguistic grounding capabilities on specific linguistic phenomena and find that VLP models struggle to ground their interdependence and relationships in visual scenes when forced to respect linguistic indicators. 
It is also suggested that more targeted investigations are needed to probe the cross-modal alignment capacity of VLP models.
Instead of the previous simple classification-based method, in this work, we introduce a new probing method based on the image captioning model, targeting for the cross-modal alignment.
As we will discuss in the following sections, our approach is able to generate sentences that enable a more transparent analysis towards the alignment of visual and textual modality.

% \section{A New Probing Task for Cross-modal Alignment Capability}

\section{Our Probing Method}
\label{sec:probing_task}

\subsection{Probing Method Overview}
\label{subsec:method_overview}

VLP models are trained to capture the relationship between different modalities.
They are able to score the matching of an image-caption pair, which can reflect whether the cross-modal semantics are aligned.
Motivated by this, in this paper we propose a new probing method to probe the cross-modal semantics alignment capability.
Specifically, we first train a captioning model to generate a caption $S$ for a given image $I$. This image-caption pair $(I, S)$ is then fed into a certain VLP model to get an alignment score $r$ (referred to as \textbf{VLP score}), which indicates the degree of alignment between images and captions from the VLP model.
Finally, the captioning model is adjusted according to this feedback score.
By continuously updating the model based on the above process, we are able to gradually generate captions with higher alignment scores from VLP models. 
We can then probe the alignment capability of VLP models, by simply evaluating the quality of these generated captions.

% A captioning model is used to generate captions expected to maximize the scores, by continuously feeding image-caption pairs into VLP models and taking the scores back.
% In this way, we are able to get the generated captions with higher scores, which are considered to fit images better and contain important alignment information which the VLP models pay more attention to.
% Thus, we can analyze these captions to understand the inner working mechanism of VLP models on cross-modal semantics alignment.

Our probing method needs to feed the scores back to the captioning model, guiding it to generate captions with higher scores. However, because of the non-differentiable problem, these scores can not be used directly to optimize the captioning model.
% \panmz{what is 'directly be fed back'? Maybe 'we cannot directly use the scores to compute gradient and update the model'?} 
To address this problem, we use self-critical sequence training (SCST; \citet{DBLP:conf/cvpr/RennieMMRG17}) , which is a standard method in image captioning task and has been widely used \citep{DBLP:conf/cvpr/00010BT0GZ18,DBLP:conf/iccv/HuangWCW19}.

\newparagraph{Self-critical Sequence Training (SCST).}
We now briefly introduce self-critical sequence training \citep{DBLP:conf/cvpr/RennieMMRG17}, a two-stage training method based on reinforcement learning. Given an image-text pair $(I, S)$, with $S=(s_0, s_1, ...  s_t)$, in the first stage, the base model tries to minimize the cross-entropy loss (CE training):
\begin{equation}\small
L(\theta)=-\sum_{t=1}^{T} \log \left(p_{\theta}\left(s_{t} | I, s_{1}, \ldots, s_{t-1}\right)\right).
\end{equation}

In the second state, SCST adopts two search strategies (greedy and sample) to generate sentences and computes the difference of a particular metric between two sentences as a reward to optimize the model. The goal of training is to minimize the negative expected reward:
\begin{equation}\small
L(\theta)=-\mathbb{E}_{S_\text{sample} \sim p_{\theta}}\left[r\left(S_\text{sample}\right) - r\left(S_\text{greedy}\right)\right],
\end{equation}
where $S_\text{sample}$ is the sample sentence, and $S_\text{greedy}$ is the greedy sentence. In \citet{DBLP:conf/cvpr/RennieMMRG17}, $r$ is a type of image captioning metric.
In our task, $r$ is the matching score of VLP models.

\subsection{Evaluated VLP Models}
\label{sec:VLP_models}

We conduct analyses of the following five VLP models, all of which have lifted the state-of-the-art results across various vision-language tasks. 
% Although these models vary in model architecture, pretraining objectives, training datasets, etc., they all lifted state of the art across many vision-language tasks. 
% Many approaches have been proposed to study these models. However, there is a lack of in-depth and targeted analysis of the cross-modal alignment of VLP models. 
% To this end, we employ VLP models to guide a simple image captioning model and investigate their respective preferred captions.
%\shi{stronger motivation}

\newparagraph{UNITER.}
\citet{DBLP:journals/corr/abs-1909-11740} propose word-region alignment via optimal transport.
% \citep{DBLP:journals/ftml/PeyreC19, DBLP:conf/icml/ChenG0LC020}. 
This task and the use of a conditional masking strategy during pretraining greatly enhance the fine-grained alignment capacity of UNITER.

\newparagraph{ROSITA.}
\citet{DBLP:conf/mm/CuiYWZZWY21} adopt an elaborate pre-training task for fine-grained alignment of different modalities. It modifies commonly-used masked language modeling and masked region modeling to structural knowledge masking, an innovative masking strategy based on the unified vision-language scene graph.
% \shi{remove MLM, not capitalize Masked Language Modeling, etc.} 

\newparagraph{ViLBERT.}
\citet{DBLP:conf/nips/LuBPL19} first introduce the two-stream architecture where image and text are encoded by two independent transformers and further fused by a co-attention mechanism. 

\newparagraph{LXMERT.}
\citet{DBLP:conf/emnlp/TanB19} also explore two-stream architecture. Compared with ViLBERT, LXMERT modifies cross-modal co-attention layers and introduces extra pre-training tasks, like ROI-feature regression and image question answering.

\newparagraph{CLIP.}
\citet{DBLP:conf/icml/RadfordKHRGASAM21} use contrastive learning to fuse visual and textual features after they are encoded separately. This simple designed task renders powerful zero-short transfer ability to CLIP across a wide range of downstream tasks, such as optical character recognition, action recognition, and text retrieval.

% \begin{table*}[h!]
% \centering
% \small
% \begin{tabular}{llllll}
% \toprule
% \textbf{Model}  & \textbf{UNITER} &  \textbf{ROSITA} &\textbf{ViLBERT} & \textbf{CLIP}  & \textbf{LXMERT} \\
% \midrule
% UNITER & 90.4$\uparrow$ & 75.4 & 89.7$\uparrow$ & 27.7 & 88.2$\uparrow$ \\
% ROSITA & 85.7$\uparrow$ & 97.3$\uparrow$ & 88.7$\uparrow$ & 26.8 & 89.1$\uparrow$ \\
% ViLBERT & 59.2 & 82.0 & 96.4$\uparrow$ & 26.2 & 74.5 \\
% CLIP & 55.0 & 86.0 & 77.0 & 32.1$\uparrow$ & 79.7 \\
% LXMERT & 73.9$\uparrow$ & 90.7$\uparrow$ & 76.1$\uparrow$ & 27.2 & 93.6$\uparrow$ \\
% \midrule
% CE & 71.6 & 86.5 & 72.5 & 27.8 & 80.1 \\
% \midrule
% Ground Truth (test) & - & - & - & - & - \\
% \bottomrule
% \end{tabular}
% \caption{Results on all VLP models. the CE model means training the FC model with cross entropy. Symbol $\uparrow$ means training with a certain VLP model has an improvement on other VLP models than using cross-entropy. }
% \label{tb:rl_results}
% \end{table*}

\begin{table*}[h!]
\centering
\small
\resizebox{0.99\textwidth}{!}{
\begin{tabular}{l|lllll|cccccc}
\toprule
\multirow{2}{*}{\textbf{Model}} & \multicolumn{5}{c|}{\textbf{VLP Model Score}} & \multicolumn{6}{c}{\textbf{Image Captioning Metrics}} \\\cmidrule{2-12}

 & \textbf{UNITER} &  \textbf{ROSITA} &\textbf{ViLBERT} & \textbf{CLIP}  & \textbf{LXMERT} & \textbf{Bleu1} & \textbf{Bleu4} & \textbf{METEOR} & \textbf{ROUGE} & \textbf{CIDEr} & \textbf{SPICE} \\
\midrule
UNITER & 90.4$\uparrow$ & 75.4 & 89.5$\uparrow$ & 27.7 & 88.2$\uparrow$ & 44.5 & 6.0 & 14.3 & 35.8 & 34.9 & 9.7  \\
ROSITA & 85.7$\uparrow$ & 97.3$\uparrow$ & 89.2$\uparrow$ & 26.8 & 89.1$\uparrow$ & 25.2 & 1.0 & 13.8 & 26.7 & 5.3 & 9.1 \\
ViLBERT & 59.8 & 80.2 & 97.3$\uparrow$ & 26.0 & 75.4 & 26.0 & 2.5 & 14.4 & 24.8 & 7.5 & 12.0 \\
CLIP & 55.0 & 86.0 & 85.5$\uparrow$ & 32.1$\uparrow$ & 79.7 & 31.2 & 3.9 & 16.4 & 31.0 & 10.3 & 10.3 \\
LXMERT & 73.9$\uparrow$ & 90.7$\uparrow$ & 84.6 & 27.2 & 93.6$\uparrow$ & 30.0 & 3.3 & 16.1 & 31.2 & 9.3 & 11.3 \\
\midrule
CE & 71.6 & 86.5 & 84.6 & 27.8 & 80.1 & 72.2 & 28.7 & 24.4 & 52.4 & 92.0 & 17.4 \\
% \midrule
% Ground Truth (test) & 92.7 & 96.7 & 85.2 & 30.3 & 94.7 \\
\bottomrule
\end{tabular}
}
\caption{Results on scores of all VLP models and image captioning metrics. CE means training the FC model with the cross-entropy loss and other models are trained under the SCST framework. Symbol $\uparrow$ means training with a certain VLP model has an improvement on other VLP models than using the cross-entropy loss. Because CLIP is different from other VLP models in measuring an image-caption pair, the scale of its score is also different.}

\label{tb:rl_results}
\end{table*}

% \begin{table*}[h!]
% % \resizebox{0.49\textwidth}{!}{
% \centering
% \small
% \begin{tabular}{lcccccc}
% \toprule
% \textbf{Model} & \textbf{Bleu1} & \textbf{Bleu4} & \textbf{METEOR} & \textbf{ROUGE} & \textbf{CIDEr} & \textbf{SPICE} \\
% \midrule
% UNITER & 44.5 & 6.0 & 14.3 & 35.8 & 34.9 & 9.7  \\
% ViLBERT & 23.8 & 2.2 & 13.5 & 23.7 & 6.6 & 12.0 \\
% CLIP & 31.2 & 3.9 & 16.4 & 31.0 & 10.3 & 10.3 \\
% ROSITA & 25.2 & 1.0 & 13.8 & 26.7 & 5.3 & 9.1 \\
% LXMERT & 30.0 & 3.3 & 16.1 & 31.2 & 9.3 & 11.3 \\
% \midrule
% CE & 72.2 & 28.7 & 24.4 & 52.4 & 92.0 & 17.4 \\
% \bottomrule
% \end{tabular}
% % }
% \caption{Results of various VLP models on image captioning metrics. Except the CE model, other models are trained under SCST framework.}
% \label{tb:rl_results}
% \end{table*}

\subsection{Experimental Setup}

\subsubsection{Training an Image Captioning Model}
\label{subsec:train_image_captioning}

\newparagraph{Dataset.} 
To train a image captioning model, we use MSCOCO dataset \citep{DBLP:conf/eccv/LinMBHPRDZ14}.
We follow the data split in \citet{DBLP:conf/cvpr/KarpathyL15} and divide the dataset into 113,287 images for training, 5,000 for validation, and 5,000 for test. 
Each image has at least five reference captions. We count all words in captions, drop the words with a frequency less than or equal to five, and finally keep 9,487 words to build a vocabulary.
In \Cref{sec:analysis} and \Cref{sec:findings}, we will use this test set for analysis. 
% \shi{complete this sent and reference to Sec 4 and Sec 5.}

\newparagraph{Captioning models.}
% \shi{change a name}
An image captioning model is essential in our probing method, as we rely on the generated captions from the image caption model to analyze the potential problems of the VLP models.
% Image captioning is a primary cross-modal task, which requires the model to automatically describe images. 
Current prevailing approaches are based on deep neural networks. 
For example, \citet{DBLP:conf/cvpr/VinyalsTBE15} feeds image features to LSTM-based language models, and various attention mechanisms are incorporated to generate better captions \citep{DBLP:conf/icml/XuBKCCSZB15,DBLP:conf/cvpr/LuXPS17,DBLP:conf/cvpr/00010BT0GZ18}. 
% Thanks to vision and language pretraining, \citet{DBLP:conf/aaai/ZhouPZHCG20} unites encoder and decoder into a single BERT-like Transformer and makes it available to fine-tune VLP models over image captioning.
% \shi{why not using BERT-like models? add in a footnote for this part.}
In this work, we use FC model \citep{DBLP:conf/cvpr/RennieMMRG17} as our image captioning model, which is similar to \citet{DBLP:conf/cvpr/VinyalsTBE15}.\footnote{We also experiment with other captioning models, such as BUTD \citep{DBLP:conf/cvpr/00010BT0GZ18} and vanilla Transformers \citep{DBLP:conf/nips/VaswaniSPUJGKP17}. In pilot studies, we observe that the constructions of generated captions are similar. We choose the FC model which has a smaller number of parameters.} 
% \citep{DBLP:journals/neco/HochreiterS97}

We train the FC model with cross-entropy loss (referred to as CE) for 30 epochs, using Adam \citep{DBLP:journals/corr/KingmaB14} optimizer with the learning rate of 5e-4. We anneal the learning rate by 0.8 every three epochs and increase the probability of feeding back a sample of the word posterior by 0.05 every five epochs. We evaluate the model on the development set every 3,000 steps and select the model with the best CIDEr score as the initialized caption model.
% \shi{or add in an advantage here.}

\subsubsection{Probing Method Setup}

\newparagraph{Probing process.}
In \Cref{subsec:train_image_captioning}, we have trained the image captioning model with cross-entropy loss (CE model).
% Now we start to probe the cross-modal alignment of VLP models.
We then follow the probing process in \Cref{subsec:method_overview} and further train this model using VLP models matching scores as rewards for extra 20 epochs and collect generated captions from all test images in MSCOCO dataset for further analysis.

% Specifically, for a given an image $I$, the captioning model generates a description $D$. Then the image-description pair $(I, D)$, is fed into a certain VLP model and the VLP model gives a matching score $r$. Note that the score in CLIP is cosine similarity, and in other VLP models is from the classifier head.
% % \panmz{how does each vlp model gives score? maybe should clarify. CLIP: cos-similarity. Others: pretrained image-text matching head.} 
% We treat the score $r$ as a reward to continue training the captioning model using the SCST framework. 
% In this way, we can get a captioning model that can generate captions conforming to a certain VLP model's preferences.

% \shi{too long.}

\newparagraph{Evaluation metrics for generated captions.}
We evaluate the quality of the generated captions using the following automatic metrics.
% Currently image captioning models are usually evaluated by automatic metrics. 
BLEU \citep{DBLP:conf/acl/PapineniRWZ02}, METEOR \citep{DBLP:conf/wmt/DenkowskiL14}, ROUGE \citep{lin2004rouge}, and CIDEr \citep{DBLP:conf/cvpr/VedantamZP15} evaluate captions based on $n$-gram overlap. 
SPICE \citep{DBLP:conf/eccv/AndersonFJG16} measures scene graph similarity between candidates and reference captions. We use the publicly released code to compute all metrics.\footnote{\url{https://github.com/tylin/coco-caption}}

\subsection{Probing Results}
\label{subsec:probing_results}

% Our experiment are all under SCST framework to probe VLP models. We compute VLP scores and common used image caption metrics for the generated sentences and results are reported in \Cref{tb:rl_results}.

We present our probing results of five VLP models in \Cref{tb:rl_results}.
% The computed VLP scores and commonly used image caption metrics for the generated sentences are reported in \Cref{tb:rl_results}.
We observe the following trends.

% \newparagraph{Each used score is improved after SCST training.} 
From the left of \Cref{tb:rl_results}, we observe that the scores on the diagonal all improve, which shows that a certain VLP score 
% \textbf{VLP score} (VLP score indicates the alignment between images and captions, and we stick to this term later) 
treated as the reward will improve after SCST training compared to previous cross-entropy training. 
It means that VLP models consider the generated sentences are more consistent with images. These results are in line with our expectations.
% \newparagraph{Using a certain score does not necessarily improve the scores of other VLP models.}
We also compute scores of generated sentences using a certain VLP score on other VLP models.
% We observe that only the score for the VLP model chosen must have gone up, while others are not necessarily.
We observe that using a certain score does not necessarily improve the scores of other VLP models. It indicates that the preferred patterns of captions of those models differ from each other.

% Moreover, there are also some interesting phenomena. 
% Using CLIP score as reward only improve CLIP score and using other VLP scores can not improve CLIP score. Using ROSITA score as reward can improve scores of all VLP models except CLIP. 

% \newparagraph{Image captioning metrics have dropped sharply.}
However, as shown in the right of \Cref{tb:rl_results}, we observe that for captions generated from our probing method, all image captioning metrics have dropped sharply compared to the CE model.
It indicates that although the captioning model can generate sentences that obtain higher scores from VLP models, which have a potential better cross-modality alignment, the quality of these sentences may not as good as we expect.
In \Cref{tb:gen_sent}, we provide three examples of the generated sentences. We observe one notable issue that captions generated by the CE model are normal and fit images, but captions after SCST training become unreadable. 
For example, the visual words in some captions are grouped together, e.g., ``\emph{elephant elephant}'', and some captions seem to follow certain sentence patterns, e.g., ``\emph{a a motor bike but a a motor but a a road}``.
% \shi{for CE?}. 
It motivates us to take a closer investigation around this abnormal phenomenon, which we will discuss in detail in \Cref{sec:analysis}.

\begin{table*}[h!]
\centering
\resizebox{0.99\textwidth}{!}{
\scriptsize
\begin{tabular}{c|m{.7\textwidth}}
\toprule
\textbf{Image} & \textbf{Generated Captions} \\
\midrule
% Example 1
\multirow{7}{*}{\begin{minipage}{.16\textwidth}\includegraphics[width=\linewidth]{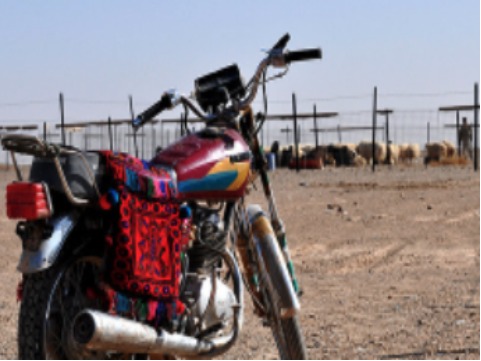}\end{minipage}} & \underline{\textbf{CE}} a motorcycle parked on a dirt road next to a fence \\\cmidrule{2-2}
& \underline{\textbf{UNITER}} a motorcycle bike motorcycle bike motorcycle bike \\
& \underline{\textbf{ROSITA}} a image that is a a motor bike but a a motor but a a road but a a dirt \\
& \underline{\textbf{ViLBERT}} a motorcycle that was in the metal at this park at the bike across the parking meter at this party \\

& \underline{\textbf{CLIP}} a image image of a row of motorcycles parked near a row area and  a wire area near a background \\
& \underline{\textbf{LXMERT}} a bike parked parked of a motorcycle parked parked of a ground near a ground near a sand sand shore\\
\midrule
% Example 2
\multirow{8}{*}{\begin{minipage}{.16\textwidth}\includegraphics[width=\linewidth]{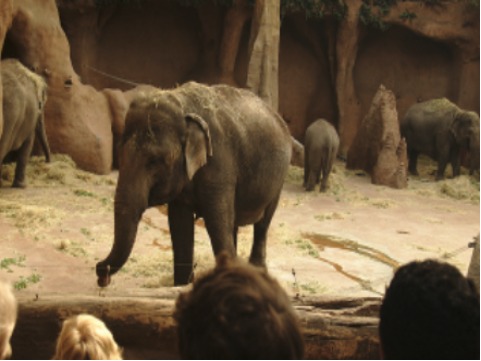}\end{minipage}} & \underline{\textbf{CE}} a group of elephants standing next to each other \\\cmidrule{2-2}
& \underline{\textbf{UNITER}} a elephant of elephants elephant \\ 
& \underline{\textbf{ROSITA}} a image image that is a a elephants elephants elephants touching a a zoo exhibit but a a zoo pin \\ 
& \underline{\textbf{ViLBERT}} a baby elephant that across the dirt towards the elephant at this enclosure behind the dirt behind his enclosure at \\
& \underline{\textbf{CLIP}} a group image of people petting a elephant near a fenced area and a baby near a background area area \\
& \underline{\textbf{LXMERT}} these elephants elephants elephants elephants interacting near a elephant elephant touching a touching a elephant near a dirt ground near\\
\midrule
% Example 3
\multirow{8}{*}{\begin{minipage}{.16\textwidth}\includegraphics[width=\linewidth]{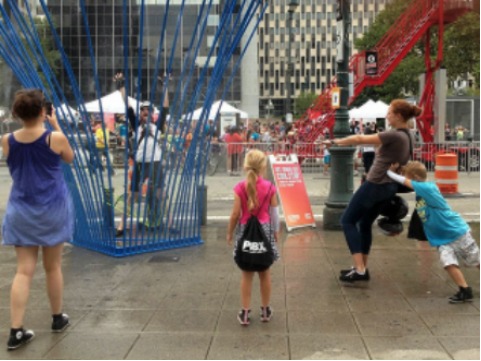}\end{minipage}} & \underline{\textbf{CE}} a group of people standing on a sidewalk \\\cmidrule{2-2}
& \underline{\textbf{UNITER}} a group playing a court playing the court \\ 
& \underline{\textbf{ROSITA}} a young people a a young person a a object that is a a court match but a a court \\ 
& \underline{\textbf{ViLBERT}} a girl that a frisbee across the basketball towards the basketball at this team at the basketball at the basketball \\
& \underline{\textbf{CLIP}} a group image of people petting a frisbee near a camera area and a neck shirt near a background area \\
& \underline{\textbf{LXMERT}} these young young people standing young standing playing a holding a walking a holding a playing a holding a sky \\

\bottomrule
\end{tabular}
}
\caption{The examples of sentences generated by captioning models trained with different VLP models. All captions are unreadable except the CE model's.}
\label{tb:gen_sent}
\end{table*}

\section{Issues of Generated Captions with High VLP Scores}
\label{sec:analysis}

In \Cref{sec:probing_task}, we have raised the issue that there exists a mismatch between increased scores from VLP models and decreased scores in standard image captioning metrics. 
In this section, we provide an in-depth and systematic analysis of these generated sentences with high VLP scores.

\subsection{Method}

We use 5,000 images from the test set of MSCOCO dataset to generate captions with high VLP scores and then perform an analysis on these sentences.
% We perform the following analyses on the generated captions with high VLP scores (after SCST training).
We first quantitatively evaluate the problem in these captions, by calculating a set of statistics at the sentence level and token level. 
For the sentence level, we count the length (Avg. Leng.) and perplexity (PPL) of captions, which reflects the influence of captions. 
The perplexity is calculated by using SRILM,\footnote{\url{https://www.speech.sri.com/projects/srilm}} a language modeling toolkit. 
We use it to train a tri-gram language model on the MSCOCO corpus. 
For the token level, we run a part-of-speech tagger to count the number of nouns.\footnote{\url{https://spacy.io}}
We count the number of nouns (Noun) and non-repeated nouns (Uni. Noun) in each sentence and the top 10 uni-grams of all captions, which can reflect the preferences of selection on tokens. 
The above statistics of generated captions are reported in \Cref{tb:sent_stats}, and the top 10 uni-grams are reported in \Cref{apdx:uni_grams}.

\begin{table}[h!]
\centering
\resizebox{0.48\textwidth}{!}{
% \begin{tabular}{p{2.5cm}lp{1.5cm}cp{1.5cm}c}
\begin{tabular}{l|cc|cc}
\toprule
\multirow{2}*{\textbf{VLP models}} &
\multicolumn{2}{c|}{\textbf{Sentence Level}} &
\multicolumn{2}{c}{\textbf{Token Level}}
% \multicolumn{2}{c}{\textbf{Sentence Pattern and Example}}\\
\\\cmidrule{2-5}
& \textbf{PPL$\downarrow$} & \textbf{Avg. Leng.} & \textbf{Noun}  & \textbf{Uni. Noun}  \\\midrule
% \textbf{\underline{Text-based metrics}} \\
UNITER & 134.6 & 6.7 & 3.2 & 2.5  \\

ROSITA & 505.2 & 19.9 & 6.6 & 4.2  \\

ViLBERT & 174.8 & 20.0 & 6.8 & 4.9  \\

CLIP & 131.2 & 19.7 & 8.7 & 6.5 \\

LXMERT & 176.6 & 19.0 & 7.2 & 3.2 \\ \midrule

CE & 7.4 & 9.5 & 3.4 & 3.2 \\ 
% Ground Truth (test) & 26.4  & 10.4 & 3.7 & 7254 \\
\bottomrule
\end{tabular}
}
\caption{
Statistics of the generated sentences. CE means training FC model with the
cross-entropy loss. At the sentence level, we calculate perplexity (PPL) and average length (Avg. Leng.).
% , where PPL represents the fluency (the lower the more fluent). 
At the token level, we calculate the average number of nouns (Noun), and the average number of non-repeating nouns (Uni. Noun). 
}

\label{tb:sent_stats}
% \vspace{-0.5cm}
\end{table}
% \makecell[c]{$DET+(NOUN)_{n}+ADP+DET+(NOUN)_{n}$ \\ a table table of the table tools}   &  1944 
% & \makecell[c]{$DET+NOUN+(AUX)_{n}+VERB+(ADP+DET+NOUN)_{n}$ \\ a man is standing on a skateboard on a sidewalk} & 352\\

\begin{figure*}[h!]
    \centering
    \includegraphics[width=0.9\textwidth]{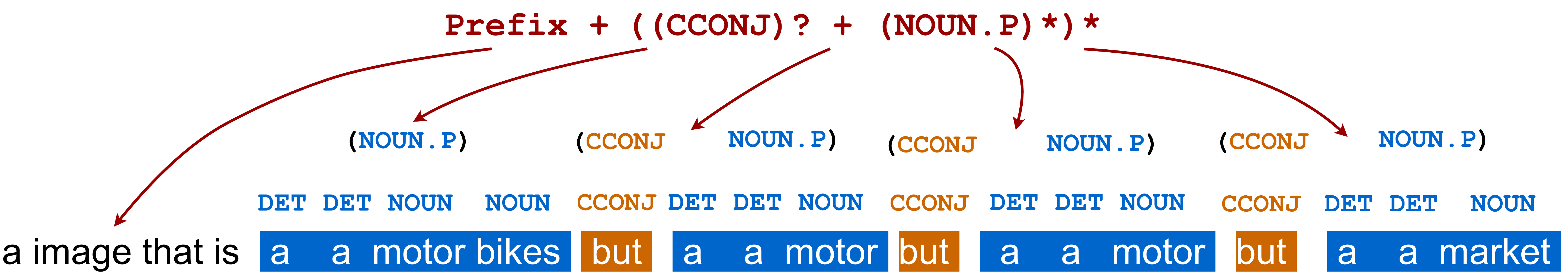}
    \caption{An example of the process of summarizing sentence patterns. The sentence is generated by the model training with ROSITA. \texttt{($\cdot$)$^*$} represents a component repeated one to more times, \texttt{($\cdot$)$?$} represents a component repeated zero or one time}
    \label{fig:parser_example}
\end{figure*}

We are also interested in the sentence patterns in the generated captions.
To do this, we first run a part-of-speech tagger for each sentence and count the $n$-grams ($n=1,2,3,4,5$). Then we summarize the patterns using regular expressions. 
\Cref{fig:parser_example} illustrates such process: Each token is first associated with a POS tag; we then extract its prefix ``\emph{a image that is}'' by $4$-grams and merge the nouns phrases \texttt{DET DET NOUN NOUN}; we finally sum up into a regular expression \texttt{Prefix+((CCONJ)?+(NOUN.P)$^*$)$^*$}.
All sentence patterns are listed in \Cref{tb:sent_pattern}.

\begin{table*}[h!]
\resizebox{0.99\textwidth}{!}{
\centering
% \begin{tabular}{p{2.5cm}lp{1.5cm}cp{1.5cm}c}
\begin{tabular}{l|cc|cc}
\toprule
\multirow{2}*{\textbf{\begin{tabular}[t]{@{}l@{}}VLP \\Models\end{tabular}}} &
\multicolumn{2}{c|}{\textbf{Prefix}} &
\multicolumn{2}{c}{\textbf{Pattern}} \\
% \multicolumn{2}{c}{\textbf{Sentence Pattern and Example}}\\
\cmidrule{2-5}
& \textbf{Top Prefix} & \textbf{Ratio} & \textbf{Top Pattern}  & \textbf{Ratio}  \\\midrule

UNITER & 
\makecell[l]{
\textbf{Prefix:} \texttt{(NOUN.P)$^*$+REL} \\ 
\textbf{Example:} a table table of} & 
\makecell[c]{ 89.7\% }  &  %4484

\makecell[l]{
\textbf{Pattern:} \texttt{Prefix+(NOUN.P)$^*$}  \\ 
\textbf{Example:} a table table of the table tools \\
} & 

\makecell[c]{87.2\% } \\ %4360
\hline

ROSITA & 
\makecell[l]{
\textbf{Pattern1:} \texttt{a+IMAGE.P+that+is} \\
\textbf{Example1:} a image image that is \\
\textbf{Pattern2:} \texttt{(NOUN.P)$^*$+that+is} \\
\textbf{Example2:} a man man that is \\
} & 
\makecell[c]{56.7\% \\ \\ 28.9\% \\ }  &  % 2833 1445 
\makecell[l]{
\textbf{Pattern:} \texttt{Prefix+((CCONJ)?+(NOUN.P)$^*$)$^*$}  \\ 
\textbf{Example1:} a image image that is a a cabinets but a a cabinets but a a sink \\
\textbf{Example2:} a man man that is a a person bike a a dirt but a a rural dirt \\ 
} & 
60.7\% \\ %3037
\hline

ViLBERT &
\makecell[l]{
\textbf{Pattern:} \texttt{(NOUN.P)$^*$+that} \\
\textbf{Example:} a bird that
} & 
\makecell[c]{91.9\%}  & % 4361
\makecell[l]{\textbf{Pattern:} \texttt{Prefix+((AUX)?+(REL)?+(NOUN.P))$^*$} \\ \textbf{Example:} a bird that the kite above the sky above the beach towards this water } & 
82.9\% \\ % 4144
\hline

CLIP &
\makecell[l]{ 
\textbf{Pattern1:} \texttt{a+IMAGE.P+of} \\
\textbf{Example1:} a image image of \\
} & 
\makecell[c]{74.6\%}  &  % 3728
\makecell[l]{
\textbf{Pattern:} \texttt{Prefix+((CCONJ)?+(REL)?+(NOUN.P)$^*$))$^*$} \\ 
\textbf{Example:} a image image of a white bathroom with a brown shower curtain near \\ a corner area and seat floor holder} & 
54.0\% \\ % 2700
\hline

LXMERT &
\makecell[l]{
\textbf{Pattern:} \texttt{(NOUN.P)$^*$+REL} \\
\textbf{Example:} a man man posing \\
} & 
\makecell[c]{99.6\%}  &  % 4981
\makecell[l]{\textbf{Pattern:} \texttt{Prefix+((REL)?+NOUN.P)$^*$} \\ \textbf{Example:} a man man posing a bike riding a bike bike riding a dirt near a road \\ near a dirt ground} & 
94.1\% \\ % 4706

\bottomrule
\end{tabular}
}
\caption{ The top prefixes and sentence patterns in generated captions. \texttt{NOUN.P} represents the nouns phrase,
\texttt{REL} represents the relationship word (including preposition and verb), \texttt{CCONJ} represents the conjunction and AUX represents the copula (including ``\emph{is}'' and ``\emph{was}''). \texttt{IMAGE.P} represents a phrase related to the word ``\emph{image}'', such as ``\emph{image image }'' and ``\emph{closeup image}''. \texttt{($\cdot$)$^*$} represents a component  repeated one to more times, \texttt{($\cdot$)?} represents a component repeated zero or one time.
}
\label{tb:sent_pattern}
\end{table*}

\subsection{Common Issues}
\label{sec:common_issues}

% There are some common issues of VLP models exposed through our probing task. Especially, 
We observe two common issues of generated captions with high matching scores, regardless of the choice of VLP models: one is these captions are not fluent, and another one is the captions contain more nouns (visual words), which are grouped together. 
% \panmz{can mention that they are the common problem of SCST framework?}
% \shi{what does grouped together mean?}

\newparagraph{Fluency issue.}
As shown in \Cref{tb:sent_stats}, generated captions using SCST training have higher perplexities (more than 100), while the perplexity is only 7.4 in cross-entropy training mode. It suggests that the generated sentences are not fluent or even unreadable (also can be seen from examples in \Cref{tb:gen_sent}). 

% It is true that these sentences have become unreadable.

\newparagraph{Noun issue.}
From \Cref{tb:sent_stats}, we observe that generated captions with higher VLP scores tend to contain more nouns (except UNITER, as its sentences are shorter). 
Also, these sentences contain many repeating nouns, while it does not happen in sentences that the CE model generates (only 0.2 difference). Sentences in \Cref{tb:gen_sent} demonstrate this abnormal phenomenon.

% In sentence level, all generated sentences' perplexities have greatly increased(more than 100), while in cross-entropy training mode, the perplexity is only 7.21. It means after SCST training, the generated sentences are far away from the dataset. Besides, these generated sentences are longer than CE. Because we set max length to 20 in inference stage, otherwise the length of the generated sentences are likely to be longer. In token level, all generated sentences contain many nouns(visual words) and few adjectives. We hypotheses it is related to training method of VLP models. VLP models are expected to align multimodality information between images and texts. But through our probing task, we show that VLP models only align objects of images and visual works of texts rather than globally align the images information and sentence semantics. In \cref{sec:va}, We further analyze the other problems of each VLP model. 

\subsection{Specific Issues}
\label{sec:va}

% In addition to the above common issues, 
We observe that VLP models may have some specific issues, which we will discuss below.
% We summarize prefixes and patterns of generated sentences in \Cref{tb:sent_pattern} and count top 10 uni-grams in \Cref{apdx:uni_grams}. 
% We find that every VLP model has its own preferences and group them into the following categories.

\newparagraph{Sentence prefix.}
We summarize prefixes of the generated captions on the left of \Cref{tb:sent_pattern}. 
We observe that ROSITA and CLIP have a special fondness for prefers related to the word ``\emph{image}''  (e.g., ``\emph{a image image that is}'' or ``\emph{a image image of}''), which accounts for 56.7\% and 74.6\% respectively.\footnote{``\emph{a image}'' is directly taken from the generated captions by the model and is not an accidental typo.}
The difference is that ROSITA prefers prefixes ending with ``\emph{of}'', but CLIP prefers prefixes ending with ``\emph{that is}''. 
We also observe prefixes that UNITER, ViLBERT, and LXMERT prefer are similar, which begin with \texttt{a+(NOUN)$^*$}. The difference is that UNITER and LXMERT favor prefixes ending with a preposition \texttt{PREP} or verb \texttt{VERB}, but ViLBERT prefers prefixes ending with ``\emph{that}''. 

\newparagraph{Sentence pattern.}
The summarized patterns of the generated sentences are shown in the right of \Cref{tb:sent_pattern}.
% We observe that each VLP model has its own favorite patterns and the model will consider sentences using these patterns are more relevant to images.
One notable observation is that all VLP models prefer to pack noun phrases together, while each VLP model has its own favorite pattern.
Single-stream architecture models (UNITER and ROSITA) hardly use a preposition or a verb to connect noun phrases. UNITER favors packing the nouns together without any connections, e.g., ``\emph{table table}''. 
ROSITA prefers to pack \texttt{determiner+determiner+NOUN} together and they are connected with ``\emph{but}'', e.g., ``\emph{a a cabinets but a a cabinets}''. 
Two-stream architecture models (ViLBERT, CLIP, and LXMERT) frequently pack \texttt{determiner+(NOUN)} together and they are connected with a preposition or verb. In particular, CLIP prefers to use ``\emph{and}'' to connect two noun phrases.

\newparagraph{Uni-Grams.}
We observe that top uni-grams of generated sentences are different between VLP models, and far away from those from model trained with cross-entropy loss (top 10 uni-grams are in \Cref{apdx:uni_grams}).
% They are far away from CE and different from each other.
We observe that single-stream architecture models pay more attention to nouns (6 in UNITER's uni-grams and 5 in ROSITA's), emphasizing on different aspects: 
UNITER focuses more aspects, e.g., animal (``\emph{cat}'', ``\emph{dog}''), place (``\emph{room}'', ``\emph{road}'' and ``\emph{apartment}'') and person (``\emph{people}''), 
while ROSITA focuses more on person (``\emph{person}'', ``\emph{man}'' and so on). 
The uni-grams of two-stream architecture models' captions are relatively uniform in types of words. 

\begin{table*}[h!]
\centering
\resizebox{0.98\textwidth}{!}{
\scriptsize
\begin{tabular}{c|m{.7\textwidth}}
\toprule
\textbf{Image} & \textbf{Constructed Captions} \\
\midrule
% % Example 1
% \multirow{4}{*}{\begin{minipage}{.16\textwidth}\includegraphics[width=\linewidth]{imgs/example1.jpg}\end{minipage}}
% & \textcolor{blue}{\underline{visual word} ``man'', ``motorcycle'', ``road''} \\
% & \\
% & \underline{\textbf{CE}} a man riding a motorcycle down a road \\
% & \underline{\textbf{Replace visual words}} a boat riding a grass down a computer \\
% & \underline{\textbf{Replace other words}} around man red a motorcycle a a road \\
% & \underline{\textbf{Replace visual words}} a boat riding a grass down a computer \\
% & \underline{\textbf{Replace other words}} around man red a motorcycle a a road \\
% image (a) & \\
% \midrule

% Example 2
\multirow{7}{*}{\begin{minipage}{.16 \textwidth}\includegraphics[width=\linewidth]{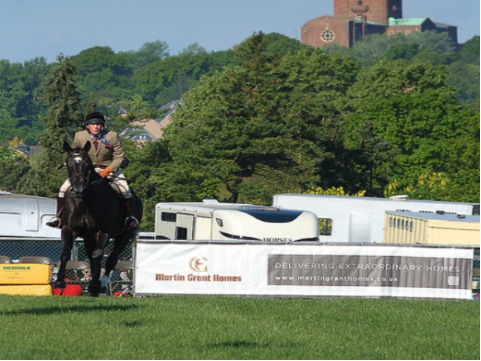}\end{minipage}}
& \underline{\textbf{CE}} a man riding a horse in a field\\
& \textcolor{blue}{\underline{Visual Words} ``man'', ``horse'', ``field''} \\\cmidrule{2-2}
& \underline{\textbf{UNITER}} a man of the horse field \\ 
& \underline{\textbf{ROSITA}} a image image that a a man but a a horse but a a field \\ 
& \underline{\textbf{ViLBERT}} a man that the horse at the field \\
& \underline{\textbf{CLIP}} a image image of a man near a horse near a field \\
& \underline{\textbf{LXMERT}} a man of a horse of a field \\
% image (a) & \\
\midrule

% Example 3
\multirow{6}{*}{\begin{minipage}{.16\textwidth}\includegraphics[width=\linewidth]{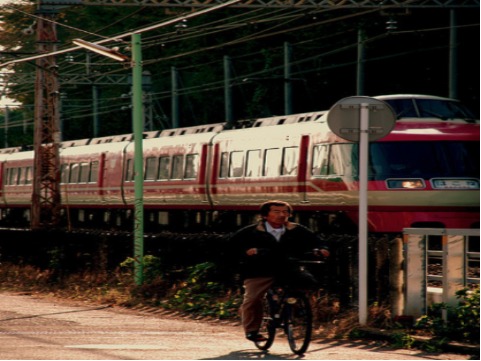}\end{minipage}}
& \textcolor{blue}{\underline{Visual Words} ``tracks'', ``person'', ``guy'', ``background'', ``bicycle'', ``man'', ``bike'', ``train''} \\
& \\
& \underline{\textbf{UNITER} } a tracks of the person guy background bicycle \\ 
& \underline{\textbf{ROSITA}} a image image that a a tracks but a a person but a a guy but a a background but a a bicycle \\ 
& \underline{\textbf{ViLBERT}} a tracks that the person at the guy at the background at the bicycle \\
& \underline{\textbf{CLIP}} a image image of a tracks near a person near a guy near a background near a bicycle\\
& \underline{\textbf{LXMERT}} a tracks of a person of a guy of a background of a bicycle  \\
% image (b) & \\
% & \\
% \midrule
% % Example 3
% \multirow{8}{*}{\begin{minipage}{.16\textwidth}\includegraphics[width=\linewidth]{imgs/img3.jpg}\end{minipage}} & \underline{\textbf{CE}} a group of people standing on a sidewalk \\
% & \underline{\textbf{UNITER}} a group playing a court playing the court \\ 
% & \underline{\textbf{ROSITA}} a young people a a young person a a object that is a a court match but a a court \\ 
% & \underline{\textbf{ViLBERT}} a child that a skateboard towards the basketball at this arena at the skate towards the sidewalk towards the crowd \\
% & \underline{\textbf{CLIP}} a group image of people petting a frisbee near a camera area and a neck shirt near a background area \\
% & \underline{\textbf{LXMERT}} these young young people standing young standing playing a holding a walking a holding a playing a holding a sky \\
\bottomrule
\end{tabular}
}
\caption{The example of constructed captions by different templates. 
(Top) exhibits the constructed captions using different sentence templates, where the visual words are from the CE caption. 
(Bottom) exhibits the constructed caption with five visual words, where the visual words are from the ground truth.}
\label{tb:attack_example}
\end{table*}
\section{Limitations of Current VLP Models}
\label{sec:findings}

The above issues reveal that VLP models tend to prefer the captions with fixed sentence patterns and more visual words.
In this section, we design a set of experiments to verify this phenomenon and further discuss the problems of current VLP models.
We argue that these limitations have potentially hindered VLP models from aligning visual and textual modalities authentically.

\subsection{Setup}
\label{sec:method}

The main idea of our verification experiments is to first construct captions, for instance to replace certain tokens, so that these captions carry characteristics that we want to test VLP models. 
These captions with their corresponding images are then fed to VLP models. 
Finally, we draw conclusions by comparing the changes in the matching scores. All experiments are done using the test set of MSCOCO dataset, containing 5,000 images.

\subsection{Findings and Discussions}

% \newparagraph{VLP models focus more on aligning objects and visual words, while ignoring global semantics.}
\begin{figure}
    \centering
    \includegraphics[width=0.48\textwidth]{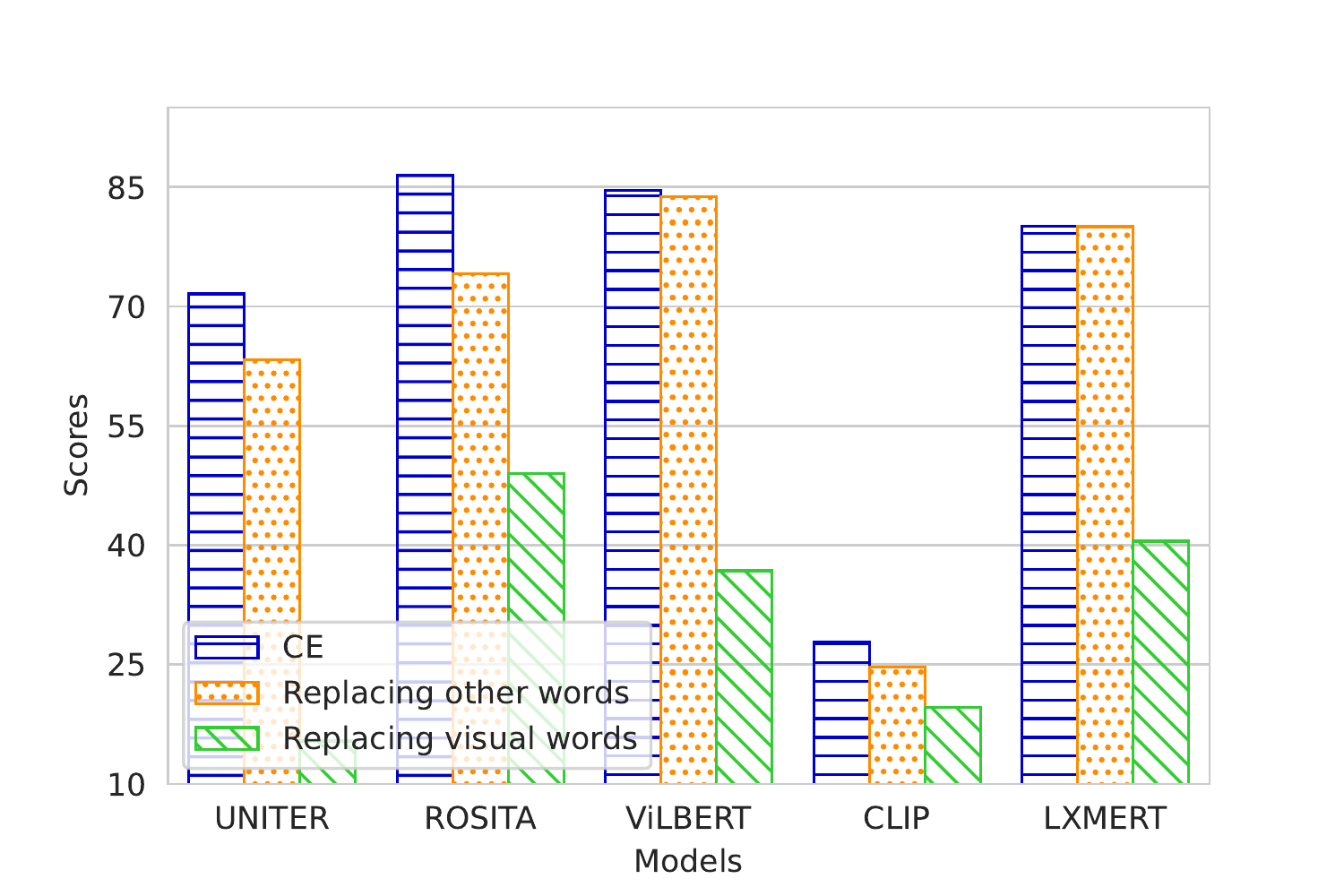}
    \caption{VLP scores of three kinds of captions. CE represents the captions generated by the model trained with cross-entropy loss. Replacing others represents the CE captions replaced all words except visual words with random words. Replacing nouns represents the CE captions replaced visual words with the wrong ones.}
    \label{fig:replace_result}
    % \vspace{-.5cm}
\end{figure}

\begin{table}[h!]
\centering
% \begin{tabular}{p{2.5cm}lp{1.5cm}cp{1.5cm}c}
\resizebox{0.49\textwidth}{!}{
\begin{tabular}{lcc|l}
\toprule 
\textbf{Model} & \textbf{CE} & \textbf{Recons.} & \textbf{Template} \\\midrule

UNITER & 71.6 & 80.5 &\texttt{a+NOUN+of+the+(NOUN)$^*$}  \\\midrule
ROSITA & 86.5 & 93.1 & \makecell[l]{\texttt{a+image+image+that+is+a+a+NOUN} \\ \texttt{(but+a+a+NOUN)$^*$}} \\\midrule
ViLBERT & 72.5 & 87.7 & \texttt{a+NOUN+that+the+NOUN+(at+the+NOUN)$^*$} \\\midrule
CLIP & 27.8 & 28.6 & \texttt{a+image+image+of+NOUN+(near+a+NOUN)$^*$} \\\midrule
LXMERT & 80.1 & 81.1 & \texttt{a+NOUN+(of+a+NOUN)$^*$}\\

\bottomrule
\end{tabular}
}
\caption{VLP model scores of reconstructed captions on various VLP models. We also present five templates that are used to reconstruct these captions.}

\label{tb:template_result}
\end{table}

\newparagraph{VLP models excessively rely on objects and visual words in their inner alignment mechanism, ignoring global semantics.}
To test if VLP models only align objects and visual words, we first prepare three kinds of captions:
(1) captions generated by the CE model (CE captions), (2) the captions that replace visual words with the wrong ones, and (3) the captions that keep the visual words but replace other words with random ones.
We then feed these captions into five VLP models and reported our results in \Cref{fig:replace_result}. 

% We use three kinds of captions designed for question (1) to test the five VLP models and reported the results in \Cref{fig:replace_result}. 
We observe that replacing visual words leads to a sharp drop in scores, but keeping them and replacing other words have a little effect on scores. 
Although the meaning of two kinds of replaced captions has changed a lot, the captions kept visual words make the VLP models deem they still match images. 
% \panmz{why those captions kept the visual words and replaced other words are far away from the origin ones?}
It shows that VLP model will directly consider a caption matches an image as long as the caption contains visual words related to the image. 
% This makes us confirm that VLP models pay more attention to aligning objects and visual words.
% \panmz{can add 'And it is difficult for VLP models to align non-visual words(verbs) with images'?}

% \newparagraph{VLP models have preferences for fixed sentence patterns.}
\newparagraph{VLP models prefer certain sentence patterns, thus ignoring more important textual information, such as fluency and grammar.}
We also experiment with reconstructing the captions based on the generated captions using cross-entropy loss, by injecting visual words into sentence templates.
We design different templates for different VLP models and list these templates in \Cref{tb:template_result}. 
We choose ``\emph{but}'' for the \texttt{CCONJ} in the template of ROSITA, because ``\emph{but}'' is the conjunction that appears most frequently in top 10 uni-grams of ROSITA. Similarly, we choose ``\emph{at}'', ``\emph{near}'', ``\emph{of}'' for the preposition in the templates of ViLBERT, CLIP and LXMERT respectively. \Cref{tb:attack_example} (Top) exhibits complete captions for these templates.

Our experimental results are in \Cref{tb:template_result}.
As we just extract visual words from the original CE model and put them in fixed sentence patterns, such reconstructed captions are normally unreadable.
However, all VLP models deem the reconstructed unreadable captions fit images better than the captions from cross-entropy loss. 
It means VLP models have preferences for fixed sentence patterns, while ignoring fluency and grammar issues.

\begin{figure}
    \centering
    \includegraphics[width=0.43\textwidth]{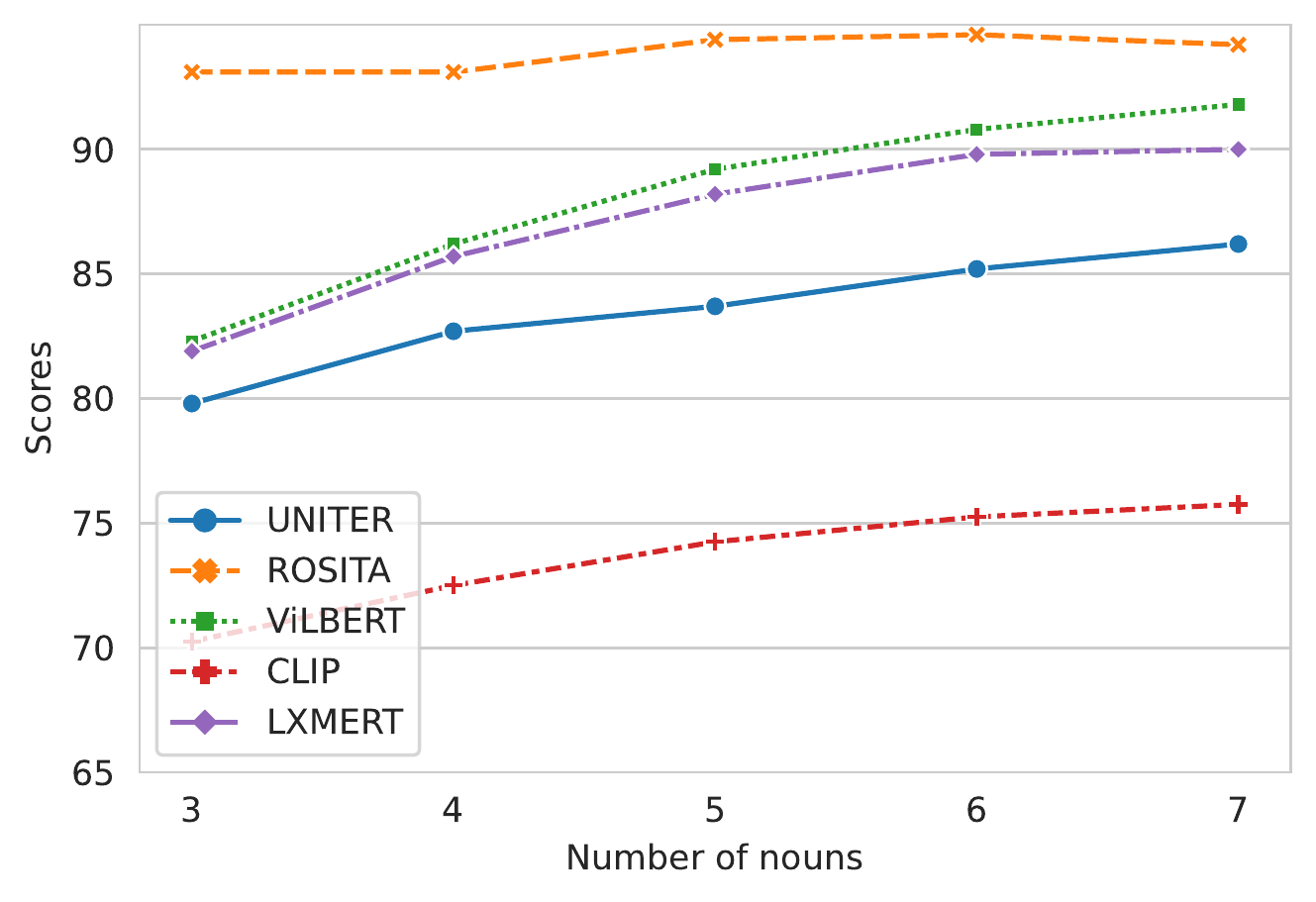}
    \caption{Scores from VLP models with varying number of nouns $k$ in the captions. For clarity, we multiply the score of CLIP by 2.5 according to \citet{DBLP:journals/corr/abs-2104-08718}.}
    \label{fig:noun_number_result}
    % \vspace{-.5cm}
\end{figure}

% \newparagraph{VLP models deem the captions containing more visual words are more consistent with images.}
\newparagraph{VLP models tend to judge captions with many visual words match images better, which might weaken the significance of key objects in images.}
We finally evaluate the role of visual tokens in VLP's cross-modal alignment.
To do this, we follow the sentence templates in \Cref{tb:template_result} to construct many captions containing visual words with different numbers. 
As the CE captions contain a few visual words (3.4 on average), we further extract visual words from the ground truth (each image with five captions at least) and merge different visual words into a set.
\Cref{tb:attack_example} (Bottom) exhibits the constructed captions with five visual words.

We vary the number of visual words $k$ ($k=3, 4, 5, 6, 7$) and report our experimental results in \Cref{fig:noun_number_result}. 
It is observed that with visual words increasing, VLP models deem the captions containing more visual words to be more consistent with images (UNITER does not change much because it gives a high score of the captions containing three visual words). It indicates VLP models deem the captions containing more visual words are more consistent with images.

% The last is that captions with many visual words tend to match images better in VLP models' judgment, which weakens the significance of key objects in images.

% \subsection{Discussions}

% The above three findings indicate three problems with the cross-modal alignment of VLP models: 
% One is that VLP models excessively rely on objects and visual words in their inner alignment mechanism, ignoring global semantics; 
% Another is that VLP models prefer certain sentence patterns, which leads them to ignore more important textual information, such as fluency and grammar;
% The last is that captions with many visual words tend to match images better in VLP models' judgment, which weakens the significance of key objects in images. We argue that these problems have potentially hindered VLP models from aligning visual and textual modalities authentically.

\section{Conclusion}

In this paper, we empirically study the cross-modal semantics alignment capability of VLP models, using a newly proposed probing method via an image captioning model. 
By analyzing the issues of generated captioning guided by the VLP models, we find that VLP models have particular weaknesses in cross-modal semantics alignment, including paying more attention to aligning objects and visual words, while neglecting global semantics; preferring fixed sentence patterns; and considering captions with more visual words are better aligned.

We hope our work sheds light on promoting better architecture or pre-training tasks for cross-modal semantics alignment to overcome these limitations. Researchers can also use our probing method to discover potential problems when designing new VLP models.

% with their models.

\section*{Limitations}

One limitation of our work is that our experiment does not cover all VLP models, as some are not open-sourced currently. 
As a result, we carefully select five VLP models that are powerful and representative of both two mainstream architectures.
We plan to experiment with more VLP models in the future to generalize our findings.

% \section*{Ethics Statement}

% May need to include a ethics statement.

% \section*{Acknowledgements}
\section*{Acknowledgements}

We would like to thank the anonymous reviewers for their constructive comments. This work was supported by NSFC No. 62176115.
\bibliography{anthology, custom}
\bibliographystyle{acl_natbib}

\newpage
\appendix
\onecolumn

\section{Top 10 Uni-Grams}
\label{apdx:uni_grams}

\begin{figure*}[h!]
    \centering
    \includegraphics[width=0.98\textwidth]{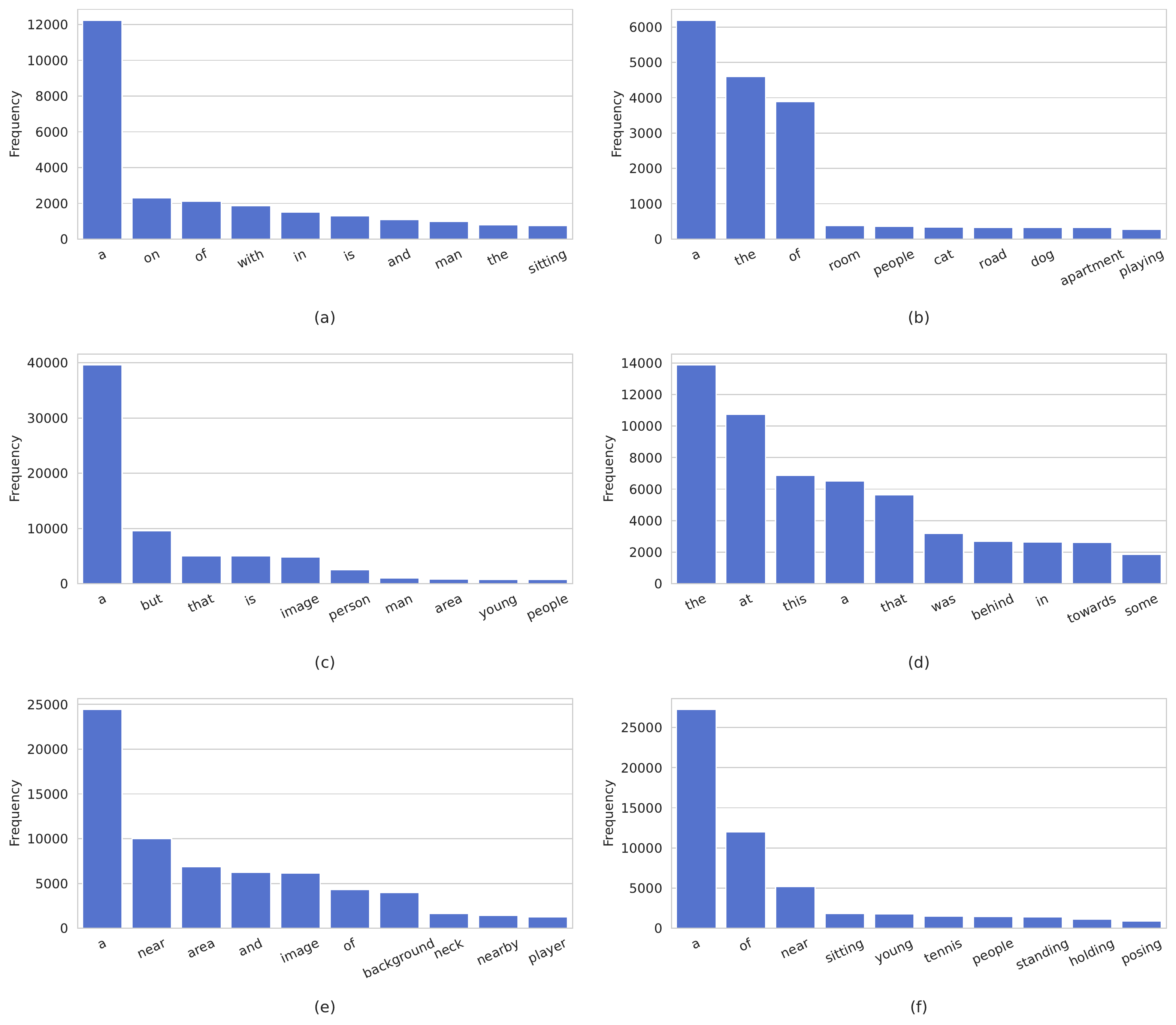}
    \caption{Top 10 uni-grams of various models. Figure (a) is from the model training with cross-entropy, and Figures (b) - (f) are from UNITER, ROSITA, ViLBERT, CLIP, and LXMERT, respectively.}
    \label{fig:grams}
\end{figure*}

We provide top 10 uni-grams of generated captions. 
\Cref{fig:grams} (b) and \Cref{fig:grams} (c) are top 10 uni-grams of UNITER and ROSITA respectively, which contain more visual words.
\Cref{fig:grams} (d),  \Cref{fig:grams} (e) and \Cref{fig:grams} (f) are top 10 uni-grams of ViLBERT, CLIP and LXMERT, which are relatively uniform in types of words. Especially, uni-grams of LXMERT' sentences contain four verbs and other models' sentences hardly have.

% \begin{figure}[h!]
%     \centering
%     \includegraphics[width=0.48\textwidth]{imgs/UNITER_one_grams.pdf}
%     \caption{top 10 uni-grams of UNITER}
%     \label{fig:uniter_grams}
% \end{figure}
% \begin{figure}[h!]
%     \centering
%     \includegraphics[width=0.48\textwidth]{imgs/ROSITA_one_grams.pdf}
%     \caption{top 10 uni-grams of ROSITA}
%     \label{fig:rosita_grams}
% \end{figure}
% \begin{figure}[h!]
%     \centering
%     \includegraphics[width=0.48\textwidth]{imgs/ViLBERT_one_grams.pdf}
%     \caption{top 10 uni-grams of ViLBERT}
%     \label{fig:vilbert_grams}
% \end{figure}
% \begin{figure}[h!]
%     \centering
%     \includegraphics[width=0.48\textwidth]{imgs/CLIP_one_grams.pdf}
%     \caption{top 10 uni-grams of CLIP}
%     \label{fig:clip_grams}
% \end{figure}
% \begin{figure}[h!]
%     \centering
%     \includegraphics[width=0.48\textwidth]{imgs/LXMERT_one_grams.pdf}
%     \caption{top 10 uni-grams of LXMERT}
%     \label{fig:lxmert_grams}
% \end{figure}
% \begin{figure}[h!]
%     \centering
%     \includegraphics[width=0.48\textwidth]{imgs/CE_one_grams.pdf}
%     \caption{top 10 uni-grams of the CE model}
%     \label{fig:lxmert_grams}
% \end{figure}
\end{document}